\documentclass[12pt,oneside,geqno]{article}
\usepackage{graphicx} 
\usepackage{amssymb}
\usepackage{amsmath}
\usepackage{amsthm}

\usepackage{natbib}
\setcitestyle{authoryear,open={(},close={)}} 

\usepackage{tikz}
\usetikzlibrary{shapes,decorations,arrows,calc,arrows.meta,fit,positioning}
\tikzset{
    -Latex,auto,node distance =1 cm and 1 cm,semithick,
    state/.style ={ellipse, draw, minimum width = 0.7 cm},
    point/.style = {circle, draw, inner sep=0.04cm,fill,node contents={}},
    bidirected/.style={Latex-Latex,dashed},
    el/.style = {inner sep=2pt, align=left, sloped}
}

\usepackage{algorithm}
\usepackage{algpseudocode}
\usepackage[english]{babel}

\theoremstyle{definition}
\newtheorem{definition}{Definition}[section]
\newtheorem*{remark}{Remark}
\newtheorem{assumption}{Assumption}[section]

\usepackage{authblk}

\begin{document}

\thispagestyle{empty}

\begin{center}
{\Large Deep Causal Behavioral Policy Learning: Applications to Healthcare}
\\[1 cm]
Jonas Knecht\footnote[1]{\scriptsize University of California, Berkeley}\\
Anna Zink\footnote[2]{\scriptsize University of Chicago} \\
Jonathan Kolstad$^{1}$\footnote[3]{\scriptsize NBER}$^*$\\
Maya Petersen$^{1}$\footnote[4]{\scriptsize University of California, San Francisco}\renewcommand{\thefootnote}{\fnsymbol{footnote}} \footnote[1]{\scriptsize Equal senior authorship}\\

\vspace*{1.0 cm}

\today

\vspace*{1.0 cm}

\end{center}

\vspace*{-1.2 cm}
\begin{abstract}
\footnotesize
We present a deep learning-based approach to studying dynamic clinical behavioral regimes in diverse non-randomized healthcare settings. Our proposed methodology - deep causal behavioral policy learning (DC-BPL) - uses deep learning algorithms to learn the distribution of high-dimensional clinical action paths, and identifies the causal link between these action paths and patient outcomes. Specifically, our approach: (1) identifies the causal effects of provider assignment on clinical outcomes;  (2) learns the distribution of clinical actions a given provider would take given evolving patient information; (3) and combines these steps to identify the optimal provider for a given patient type and emulate that provider's care decisions. Underlying this strategy, we train a large clinical behavioral model (LCBM) on electronic health records data using a transformer architecture, and demonstrate its ability to estimate clinical behavioral policies. We propose a novel interpretation of a behavioral policy learned using the LCBM: that it is an efficient encoding of complex, often implicit, knowledge used to treat a patient. This allows us to learn a space of policies that are critical to a wide range of  healthcare applications, in which the vast majority of clinical knowledge is acquired tacitly through years of practice and only a tiny fraction of information relevant to patient care is written down (e.g. in textbooks, studies or standardized guidelines).

\end{abstract}

{ \small
\begin{quote}
\textbf{Keywords:} \\
\textbf{JEL Classification:}  
\end{quote}
}

\clearpage

\renewcommand{\thefootnote}{\arabic{footnote}} 

\setcounter{page}{1}

\section{Introduction}
Healthcare provision requires individual human decision makers to act in settings of substantial uncertainty and decision complexity where the majority of patient care exists in the gray area of medicine, without clear guidelines or evidence from clinical trials. As a result, healthcare is characterized by widespread variation in clinical practice and associated differences in outcomes and costs, for otherwise identical patients (e.g. \cite{glover1938incidence}, \cite{finkelstein2016sources}). Mitigating such differences through the more effective determination of optimal patient-specific treatments has the potential to transform healthcare delivery by driving better health and lower cost. Doing so, however, requires capturing clinical knowledge on a granular level (i.e. individual patient by action) and deploying it at scale (i.e. across the entire healthcare system). 
\\
\\
Capturing real-world clinical knowledge at scale remains one of the fundamental challenges in healthcare. Traditional solutions, such as rule-based engines, require extensive manual effort and are difficult to maintain, while newer approaches, such as large language models (LLMs), grapple with hallucinations and struggle to predict real-world clinical actions (see \cite{Hager_2024_LLM_MedLimitations}). Language-based approaches are inherently limited by their reliance on written information sources. They are, by construction, ''textbook" medicine. Any solution that relies solely on language will fall short since, particularly in healthcare, actions are the primary repository of knowledge and the complex reasoning processes learned by providers over years of clinical practice are nowhere written down at scale. 
\\
\\
In this paper, we combine advances from the causal inference, statistics, economics, and computer science literature to propose a solution that addresses these challenges. At the heart of our methodology is a fundamental tenant of economics: that observed human behavior encodes complex and often implicit or tacit knowledge (\cite{samuelson1938consumer}, \cite{Polanyi1966}). If we can learn the actions taken by clinicians, we can access the knowledge that providers accumulate over years of treating real-world patients. 
\\
\\
Our approach leverages recent advances in deep learning to model provider decision making in response to evolving clinical information as a sequence-to-sequence learning task. We call this model a \textbf{large clinical behavioral model}, or \textbf{LCBM}, as it learns to emulate the behavior of clinicians choosing actions for their patients at scale. At the same time, we apply methods from semiparametric efficient estimation of causal effects to causally identify high-quality providers, i.e. providers who achieve the best expected patient outcomes. Finally, building on machine learning approaches for optimal regimes and policy learning, we integrate this information with the LCBM to learn how top providers would have treated a given patient. We refer to this process as \textbf{deep causal behavioral policy learning}. Its objective is to learn the clinical reasoning processes employed by a range of providers, and then to hone in on the practice patterns of the providers who are causally linked to improved patient outcomes, and thus to provide high quality clinical reasoning at scale. Achieving this requires the interaction of the aforementioned disciplines as we present a systematic framework for both identifying high quality clinical practice and capturing this with modern deep learning methods. 
\\
\\
Our proposed methodology supports a range of practical clinical applications, including causally rigorous quality measurement, provider coaching, and causally-grounded clinical decision support for complex longitudinal care, as well as a tool for tuning clinical reasoning models. 

\subsection{Paper Overview}

We start by defining a general Structural Causal Model (\cite{Pearl_2009}) to describe the causal processes that generate clinician behavior and patient states in healthcare settings. More concretely, the model encodes the following data-generating process: After a patient is assigned to a given provider, the provider makes a decision about which clinical actions to take based on information on that patient at that time point. Clinical actions taken by the provider such as laboratory or imaging tests ordered reveal additional information about the patient. This process is repeated recursively over the course of a patient's care journey, which may span multiple encounters depending on the clinical setting. We refer to this as a dynamic longitudinal \textit{behavioral} policy. 
\\
\\
Next, we define the \textbf{optimal provider} as the provider which, for a given patient type, is able to achieve the best counterfactual patient outcomes. Our core identification assumption is that provider assignment is random conditional on the full set of observables at time of provider assignment. This assumption, combined with an assumption of sufficient provider-specific data support, allows us to identify the causal impact of provider assignment on patient outcomes. Using the same assumptions, we can identify the counterfactual behavioral policy of the optimal provider for a given provider information-set., which we call the \textbf{optimal behavioral policy}.
\\
\\
Our estimation tasks are characterized by a high dimensional, non-parametric statistical model. This introduces substantial challenges, particularly when estimating the optimal provider (in a large set of candidate providers) for a given information set. In order to estimate such quantities, we rely on advances in semiparametric efficient estimation of causal effects and machine learning approaches for optimal regimes. Crucially, these estimators allow for the use of deep-learning methods, which are essential given the dimensionality of the data used for estimation. 
\\
\\
To learn a provider-specific behavioral policy, we propose a simple two step procedure in which one first pre-trains a behavioral policy estimator, based on a transformer architecture, on clinical action path data from the entire population of providers and patients, before selectively fine-tuning towards the practice patterns of particular providers. Similar methods, alongside more complicated reinforcement learning-based algorithms (the focus of ongoing work), have increased in popularity, especially with transformer-based large language models (\cite{RLHF}, \cite{LoRA}, \cite{Distillation}, \cite{GPT}, \cite{BERT}). Through this process one can generate a series of provider-specific behavioral policy estimators. Combining this with the optimal provider assignment mechanism discussed above, we can estimate the counterfactual behavioral policy of the optimal provider for a given patient-type. We call this procedure \textbf{deep causal behavioral policy learning}. 

\subsection{Overview of Relevant Literature}
Our core identification strategy builds on a rich literature that uses providers as (conditional) instrumental variables for the causal effects of treatment decisions (\cite{Doyle_2015}, \cite{Chan_2022}, \cite{Kling_2006}, \cite{Smulowitz_2021}, \cite{Brookhart_2007}, \cite{Brookhart_2006}, \cite{Wang_2005}, \cite{Korn_1998}).  We use comprehensive large multi-provider electronic health record datasets to improve the plausibility of these assumptions, and employ semiparametric efficient estimation approaches (see, e.g. \cite{Laan_Robins_MethodsCensoredData}, \cite{Laan_Rose_TargetedLearning}, \cite{Chernozhukov_DRML}, \cite{AtheyWagner2021}) that permit the integration of machine learning in order to fully leverage the information in these datasets.  In particular, we draw on the rich literature on machine learning methods for learning individualized treatment rules (or ``optimal dynamic regimes") and the value of these regimes ( e.g., \cite{Kosorok_Laber_PrecisionMed}, \cite{Chakraborty_Moodie_DTR}, \cite{Luedtke_2016}, \cite{Luedtke_Laan_SuperLearning_ODTR}, \cite{Luedtke_Laan_TargetedMeanOutcomeODTR}, \cite{Athey_Imbens_2016}, \cite{AtheyWagner2021}). Recent related work has focused on the use of instrumental variable methods for optimal dynamic regime estimation (\cite{Pu_2021}, \cite{Cui_2020}, \cite{Qiu_ReJoinder_2021}, \cite{Han_2020}). We differ from these methods in our target of estimation (estimand), and our use of providers as multi-dimensional categorical instruments.
\\
\\
Like our work, others have proposed methodology premised on the insight that some provider behavior may already encode an optimal (or near optimal) treatment policy, and the utility of leveraging the information encoded in these provider choices in settings where treatment effects are confounded (\cite{Stensrud_2024}, \cite{Luckett_2021} \cite{Wallace_Moodie_2018}, \cite{Pu_2021}). We use recent advances in deep learning and transformers (\cite{SeqToSeq}, \cite{AttentionIsAl}) to encode provider knowledge more effectively, and outline the ability to combine this with machine learning approaches to provide individualized matches of patients to skilled providers. We demonstrate the ability of even relatively simple transformer models to learn the complex relationship between a patient's history of clinical actions and the likely next step chosen by a given provider. We add to recent work (\cite{EHRMamba}, \cite{ETHOS}, \cite{MOTOR}) that also includes the use of transformers, and variations thereof, to estimate the conditional distribution of longitudinal clinical action paths for downstream outcome prediction. Our work differs from this in our framing of causal questions, formal causal identification strategy, and our use of transformers, coupled with fine-tuning, to estimate the behavioral policies of particular providers. As we describe, this opens the door to a range of downstream applications, including new approaches to clinical decision support, quantifying causal variation in outcomes, and training clinical reasoning models.

\subsection{Paper Outline}

The remainder of the paper is organized as follows. Section 2 establishes our formal causal model and identification results. Here, we outline identification of the optimal provider and optimal behavioral policy. Section 3 continues with estimation and presents potential estimators for the expected counterfactual patient outcome under alternative provider assignment policies, for the optimal provider, and for the optimal behavioral policy. In particular, we review why transformers are particularly well-suited for behavioral policy learning, and propose embedding and training approaches. Section 4 presents preliminary empirical results from an LCBM pre-trained using a real-world EHR data from the UCSF emergency department. We conclude with a discussion of limitations, open questions, methodological extensions, and clinical applications of this highly general methodology. 

\section{Causal Model and Identification Results}
To simplify exposition, we consider the problem of identifying an optimal provider at a single time point for a given patient (a point treatment causal inference problem). Specifically, we focus on a setting in which a clinical provider is assigned at a given time point (e.g., the start of an encounter) and is primarily responsible for subsequent care delivery decisions within a given clinical domain until a clinical outcome is measured.  

\subsection{Observed data}
We consider the following longitudinal data structure at the patient level. At a given time point $t$, for a given patient $i$, we measure patient characteristics, $X_{i,t}$. These characteristics, or states,  can include a wide range of individual patient medical histories (e.g., diagnosis codes, laboratory values, medical images, and clinical notes), other patient measures such as social determinants of health, location of residence, and characteristics of the setting in which care is delivered (e.g., facility size, characteristics of the patient population served). We further measure a patient's provider(s), $J_{i,t}$, and the set of clinical actions ordered for that patient $A_{i,t}$, which can include a wide range of treatments, interventions, and monitoring decisions.   
\\
\\
Let $t=0$ denote the first time point that a patient appears in the database. We assume an arbitrarily small discrete time scale corresponding to the frequency of actions and state changes (noting that the duration of incremental time intervals will vary by clinical setting), with time ordering in a given time interval of $(X_{i,t},J_{i,t},A_{i,t})$. Throughout, we use notation $\boldsymbol{Z}_{0:t}\equiv (Z_{t=0},...,Z_{t})$ to denote the longitudinal history of a random variable $Z$ through time $t$. Patients are observed in the dataset up to a maximum time point $T$, with the full observation interval $t\in [0,T]$ potentially spanning multiple patient encounters. Let $K$ denote the time point at which the provider assignment of interest occurs, and let $J_K$  denote the provider assigned at that time point; in slight abuse of notation, we sometimes use $J \equiv J_K$. We focus on the case where the same provider is responsible for the care decisions studied from assignment until the outcome is measured, i.e., $\boldsymbol{J}_{K+1:T}=J_K$. 
\\
\\
We define a clinical outcome of interest that occurs after provider assignment. For ease of exposition, we focus on a single outcome  $Y \equiv Y_T\in X_T$, assessed completely at the end of follow-up $t=T$. (Our methodology generalized naturally to a wide range of alternative outcome types, including multivariate and time-to-event outcomes, and ``intermediate" outcomes measured before time $T$). Our notation is compatible with encounter-level analyses, and our choice of time-indexing is fully general. Without loss of generality, we assume that larger values of $Y$ indicate better outcomes. The total observed data on a random patient thus consist of;
$$
\boldsymbol{O}= (\boldsymbol{X}_{0:T},\boldsymbol{J}_{0:T},\boldsymbol{A}_{0:T}).
$$
Let $P_0$ denote the distribution of $\boldsymbol{O}$.  Our objective of identifying the provider assignment policy at time $K$ and the corresponding optimal behavioral policies from $K$ to $T$ that will optimize the outcome $Y$ motivates a particular factorization of this distribution: 
\begin{equation}
    P_0=Q_{0,I_J}g_0\boldsymbol{\pi_0}\boldsymbol{\Xi_{0}}
\end{equation}
where;
\begin{itemize}
    \item $Q_{0,I_J}$ denotes the distribution of the ``provider information set" $I_J \equiv (\boldsymbol{X}_{0:K},\boldsymbol{J}_{0:K-1}, \boldsymbol{A}_{0:K-1}) \in \mathcal{I_J}$, corresponding to the full observed history of a patient prior to provider assignment at time $K$. 
    \item $g_0\equiv g_0(J|I_J)$ is the ``provider mechanism", i.e., the conditional distribution of provider assignment at time $t = K$ given the provider information set. Note that this is the basis for a provider-level propensity score; however,  it differs from commonly used binary (or low-dimensional) propensity scores since it maps information sets $I_J$ to probability distributions over a  large space of potential providers $J \in \mathcal{J}$ where $|\mathcal{J}| >> 2$. 
    \item $\boldsymbol{\pi}_{0} \equiv   
    \prod_{t = K}^{T} \pi_t(A_t \; | \; \boldsymbol{I}_{0:t})$ is the ``action mechanism" or ``behavioral policy" from time of provider assignment until the outcome is measured,  where  $\boldsymbol{I}_{0:t} \equiv (\boldsymbol{X}_{0:t}, \boldsymbol{J}_{0:K},\boldsymbol{A}_{0:t-1}) \in \mathcal{I}$ is the information set\footnote{Note that we sometimes use $I_{t}$ instead, but unless otherwise stated we always refer to a patient's complete history.} (i.e. the full observed history) available just before choice of actions $A_t$ at time $t$, and $\pi_{t}(A_t \; | \; \boldsymbol{I}_{0:t})$ represents the observed behavioral action policy - i.e., the information-set conditional probability distribution over the space of potential actions at each period $t$. Note that conditioning on $J_K = j$ yields a provider-specific behavioral policy (i.e., the observed action policy of a provider $j$), which we denote as $\boldsymbol{\pi}_{0}^{j}$.
    \item $\boldsymbol{\Xi_0} \equiv \prod_{t=K+1}^T F_{t}(X_t \; | \; \boldsymbol{X}_{0:t-1}, \boldsymbol{J}_{0:K},\boldsymbol{A}_{0:t-1})$  is the conditional distribution of observed patient characteristics given the observed past, from time of provider assignment until the outcome is measured (recalling that the outcome $Y \in X_T$). 
\end{itemize}

\subsection{Causal Model}
We  assume a structural causal model $\mathcal{M^F}$ on the  data generating process for this observed data structure, with endogenous nodes $\boldsymbol{O}$, and corresponding exogenous nodes (latent errors) $\boldsymbol{U}\equiv ((U_{X_{t}},U_{J_{t}},U_{A_{t}}): t=0,...,T)$. Denote the true unknown joint distribution of $(\boldsymbol{O},\boldsymbol{U})$ as $P_{\boldsymbol{O},\boldsymbol{U}} \in \mathcal{M^F}$. The endogenous nodes are covered by two separate data manifolds corresponding to the observed data, which cover our patient features $\boldsymbol{X}$ (including the outcome $Y$), time-varying provider action paths $\boldsymbol{A}$, and provider assignments $\boldsymbol{J}$ (including assignment of the provider of interest, $J_K$). Our casual model on the full data generating process thus implies a  statistical model on the the set of possible observed data distributions; denote this statistical model $\mathcal{M}$, such that $\boldsymbol{O}\sim P_0 \in \mathcal{M}$.
\\
\\
A series of structural equations represent the causal links between different nodes in the implied DAG shown in figure 1. We assume a simple individual-level structural causal model (\cite{Pearl_2009}), which encodes no interference or spillover. Here we make use of the differentiated latent features $((U_{X_t},U_{J_t},U_{A_t}): t=0,...,T)$ and let $f$ represent the particular structural form. This gives rise to the model below: 
\begin{align}
X_t &=
    \begin{cases}
            f_{X_t}(\boldsymbol{X}_{0:t-1}, \boldsymbol{J}_{0:t-1},\boldsymbol{A}_{0:t-1},U_{X_t}), & t \in \{0,K\},\\
            f'_{X_t}(\boldsymbol{X}_{0:t-1},\boldsymbol{J}_{0:K-1},\boldsymbol{A}_{0:t-1},U_{X_{t}}), & t \in \{K+1,T\}
    \end{cases}\\
    \\
    J_t &= f_{J_t}(\boldsymbol{X}_{0:t}, \boldsymbol{J}_{0:t-1}, \boldsymbol{A}_{0:t-1}, U_{J_t}),t\in \{0,K\} \\
    A_{t} &= f_{A_t}(\boldsymbol{X}_{0:t}, \boldsymbol{J}_{0:t}, \boldsymbol{A}_{0:t-1}, U_{A_t}) t\in \{0,T\}
\end{align}
For notational convenience, we define any time-varying vector indexed by ${0:-1}$ as the empty set $\{\}$. We place no restrictions on the functional form of the corresponding structural equations. We  highlight two important assumptions we make throughout. 
\begin{assumption}(Exclusion Restriction)
    We assume that $J_K$ can impact $X_{K:T}$ only through $A_{K:T}$, i.e., providers can impact patient states (and thus the outcome) only through recorded actions. We assume no other exclusion restrictions and that each endogenous node may be affected by any of the factors that precede it.
\end{assumption}
\begin{assumption}(Independence Assumption)
    We assume that latent factors determining $J_K$ are independent from the latent factors for all subsequent nodes in our causal model $\mathcal{M}^F$ such that: 
    \begin{equation}
        U_{J_K} \perp (U_{I_J}, U_{\boldsymbol{X}_{K+1,T}},U_{\boldsymbol{A}_{K,T}}),
    \end{equation}
\end{assumption}
\noindent
(where we use $U_{I_J}$ to refer to the latent factors for all elements of the provider information set $I_J$). Taken together, these assumptions imply that the provider of interest $J_K\equiv J$ is an instrument (conditional on observed information) for the effect of an arbitrarily complex behavioral policy on the outcome of interest $Y$. The resulting casual model can be visualized as the (slightly simplified) directed acyclic graph (DAG)  in Figure 1.
\begin{center}
            \vspace{0.3cm}
            \begin{tikzpicture}
            
                \node[state, circle] (I_0) at (0,0) {$I_{J}$};
                \node[state, circle] (U) at (3,0) {$U$};
                \node[state, circle] (U_J) at (-2,-2) {$U_{J}$};
                
                \node[state, circle] (J) at (0,-2) {$J$};
                \node (box_point) at (3, -1.55) {};
                
                \node (A_t) at (2,-2) {$A_K \; $};
                \node (X_t) at (2,-3.5) {$X_{K+1} \; $};
                \node (A_t+1) at (4,-2) {$A_{K+1+n}$};
                \node (X_t+1) at (4,-3.5) {$X_{K+1+n}$};

                \node[state, circle] (Y) at (7,-2.74) {$Y$};

                \path (I_0) edge (J); 
                \path (I_0) edge[bend left=12]  (Y); 
                \path (I_0) edge[bend left=0] (box_point); 

                \path (J) edge (A_t); 
                \path (A_t) edge (X_t+1); 
                \path (A_t) edge (A_t+1);  
                \path (X_t) edge (X_t+1); 
                \path (A_t) edge (X_t); 
                \path (X_t) edge (A_t+1); 
                \path (X_t+1) edge (A_t+1); 
                \path (U_J) edge (J); 

                \path (U) edge (I_0); 
                \path (U) edge (box_point); 
                \path (U) edge[bend left=25]  (Y); 
                \node[draw=black, fit=(A_t) (X_t) (A_t+1) (X_t+1)] (box) {};
                \path (box) edge (Y); 
            \end{tikzpicture} \\
            \text{\textbf{Figure 1:} Time-varying DAG, $n \in \{1,T-1-K\}$} 
\end{center} 
\noindent
\\
Importantly, our causal model allows for an arbitrarily complex set of interactions between the time-series of patient states $X_{0:T}$ and physician actions $A_{0:T}$,  allowing in particular for arbitrarily complex observed behavioral policies implied by $\boldsymbol{\pi}_0$. As we outline in Section 3, our proposed methodology leverages modern deep learning methods with high degrees of data adaptivity to learn these policies given extremely high-dimensional information sets and action spaces. 

\subsection{Causal Estimands and Identification Results} 
Having outlined our observed data structure as well as our causal model, we now specify a series of the causal estimands and provide corresponding identification results. Specifically, we define and identify the impact of hypothetical provider assignment on expected patient outcome, the implied optimal provider assignment policy, and the counterfactual behavioral policy of a hypothetical provider.

\subsubsection{Optimal provider policy}
First, we consider the problem of identifying the optimal provider for a given patient information set, as well as the value of such a provider assignment policy (i.e., the expected counterfactual outcome if patients were assigned to providers according to such an optimal policy). For a given outcome, the difference between the expected counterfactual outcome under the hypothetical optimal provider assignment policy and the expected observed outcome provides a summary measure of care quality, in that it summarizes, for a given target population of patients or clinical care setting, the net impact of suboptimal actions and action paths on patient outcomes. 
\\
\\
Consider a hypothetical intervention to assign a patient to provider $j \in \cal{J}$. Let $d_J(I_J)$ denote a policy that assigns a hypothetical provider $j \in \cal{J}$ based on observed information set $I_J$ (i.e., $d_J$ is a function that maps an information set $I_J$ to a provider $j \in \mathcal{J}$, with special case corresponding to assignment of a single provider $d_J(I_J)=j$). Let $\mathcal{D}$ denote the set of candidate provider assignment policies. Note that, in practice, one may modify the information set $I_J$, provider set $\cal{J}$, or set of candidate policies $\mathcal{D}$ to reflect the setting-specific context or to improve the plausibility of identification assumptions. Let $Y^{d_J(I_J)}$ (sometimes abbreviated $Y^{d_J}$) denote a patient's counterfactual outcome under a candidate provider assignment policy $d_J(I_J)$ that assigns each patient to the provider indicated by policy $d_J(I_J)$. A model on the distribution of the counterfactual random variable $Y^{d_J(I)}$ is defined based on a hypothetical intervention to set $f_{J_K}=d_J(I_J)$ on the causal model $\mathcal{M^F}$. 
\\
\\
We can now define the \textbf{optimal provider policy}, denoted $d^*_J$, as: 
\begin{equation}
    d^*_J\equiv \arg\max_{d_J(I_J)}\mathbb{E}_{P_{\boldsymbol{O,U}}}(Y^{d_J}).
\end{equation}
Identification of the optimal provider policy  follows from Assumption 2.2 (which implies $ Y^{d_J} \perp J | I_J$) together with an assumption of ``positivity," i.e., sufficient data support for the range of possible providers to whom a patient might be hypothetically assigned given the value of the information set (\cite{Petersen_2012}).  We can ensure that the positivity assumption holds by design by restricting the set of candidate providers considered (either overall, or for a given information set) to those with sufficient support (see \cite{Laan_Petersen_2007}). 
\\
\\
Let $\bar{Q}_0(j, I_J) \equiv \mathbb{E}_0(Y|J = j, I_J)$ for $j \in \mathcal{J}$ where $\mathcal{J}$ is the potential set of providers. Then, under the assumption above we can identify the expected counterfactual patient outcome under a candidate provider assignment policy as; 
\begin{equation}
    \mathbb{E}_0(Y^{d_J}) = \mathbb{E}_{0}(\bar{Q}_0(J = d_J(I_J), I_J)), 
\end{equation}
where the right hand side is a function of the observed data distribution $P_0$. By extension, the optimal provider policy, and the value of the optimal provider policy are also identified as parameters of $P_0$. Alternative identification strategies at this stage are also possible.

\subsubsection{Provider behavioral policy}
We next define and identify provider-specific counterfactual behavioral policies. Let  $\boldsymbol{\pi}^{j}$ denote the counterfactual behavioral policy of a specific provider $j$ that captures the mapping from time-updated information sets to actions that would be taken by that particular provider over time:  $\boldsymbol{\pi}^{j} \equiv \prod_{t=K}^T \pi^j_t(A_t^j|I_J,\boldsymbol{X}_{K+1:t}^j, \boldsymbol{A}_{K:t-1}^j)$, where again we use a superscript $j$ to denote a counterfactual random variable or distribution under a hypothetical intervention to assign provider $j$ (and where for $t=K$ we define $\boldsymbol{A}_{K:K-1}^j \equiv \{\}$ and $\boldsymbol{X}_{K+1:K}^j \equiv \{\}$). the same assumptions (Assumption 2.2 together with sufficient data support) allow us to identify the counterfactual distribution of action paths taken by any given provider (and by extension, the optimal provider $d^*_J$, as defined above) using the g-computation formula (\cite{Robins_1986}). Specifically, the counterfactual provider-specific behavioral policy under hypothetical provider assignment $j$ is equivalent to evaluating the observed behavioral policy $\boldsymbol{\pi}_{0}$ at $J = j$:
\begin{equation}
  \boldsymbol{\pi}_0^{j} = \prod_{t=K}^T \pi_{0,t}(A_t|I_J, J=j, \boldsymbol{X}_{K+1:t}, \boldsymbol{A}_{K:t-1})
\end{equation}
A specific instance of this is $\boldsymbol{\pi}_{0}^{d^*_J}$, which we call the optimal behavioral policy and denote by $\boldsymbol{\pi}_{0}^*$.
\begin{definition}[Optimal provider-specific behavioral policy]
    A provider-specific behavioral policy $\boldsymbol{\pi}_{0}^j$ that captures the behavior of decision maker $j$ is optimal for a patient with information set $I_J$ when $j = d^*_J(I_{J})$. We define the \textbf{optimal behavioral policy} as; 
    \begin{equation*}
        \boldsymbol{\pi}_{0}^* \equiv \boldsymbol{\pi}_{0}^{d^*_J(I_J)}
    \end{equation*}
\end{definition}
\noindent
That is, the policy is  optimal  when it has learned the behavior of the optimal counterfactual physician assignment. Optimality here is defined with respect to the outcome $Y$ used to construct $d^*_J$.  

\subsubsection{Connecting $\boldsymbol{\pi}$ and $d^*_J$}
In this section we aim to bring together the motivation behind the estimands outlined above, i.e., how do we aim to utilize $d^*_J$ and $\boldsymbol{\pi}_0$. We begin by summarizing the main points from the sections above;  
\begin{enumerate}
    \item Under our assumption of conditional random assignment of providers we are able to identify the expected counterfactual outcome $\mathbb{E}_{P_{O,U}}(Y^{j})$ for assignment to any $j$.
    \item By extension, we are able to identify the optimal provider $d^*_J(I_J)$ conditional on $I_J$.
    \item Under the same assumptions, we are able to identify the counterfactual action policy $\boldsymbol{\pi}^{j}_0$ for any $j$.
    \item By extension of the above, we are able to identify the optimal counterfactual action policy $\boldsymbol{\pi}^*_0$. 
    \item Together, this also allows us to identify the expected counterfactual outcome $\mathbb{E}_{P_{O,U}}(Y^{d^*_J})$ had each patient been assigned their optimal provider (given $I_J$), and the corresponding counterfactual behavioral policies $\boldsymbol{\pi}^{j}_0$ of these optimal providers. 
\end{enumerate}
Other approaches in this literature have often focused on directly identifying the impact of actions (and series of actions) on patient outcomes, where in real-world data settings in which actions are not randomly assigned, identification relies on strong assumptions about absence of unmeasured confounders (unmeasured shared common causes of actions and outcomes). With the approach proposed here, we utilize an alternative identification approach, focusing on quasi-random assignment of providers (i.e. no unmeasured confounding of provider assignment) rather than quasi-random treatment of patients. In many clinical settings, such as the emergency department, this is often plausible. In its own right, this approach provides a rigorous quality assessment, allowing us to identify the ``value" of providers, i.e., the expected outcomes any given provider would have achieved in the full population $\mathbb{E}_{P_{O,U}(Y^{j})}$, or a subset of the population, as well as the maximum expected counterfactual outcomes achievable under optimal provider assignment $d^*_J(I_J)$. This allows us to identify the distribution of provider ``skill", and provide benchmarks for quality assessment. 
\\
\\
Furthermore, under Assumption 2.1 that provider assignment affects clinical outcome only through effects on measured actions (i.e. by changing behavioral policy), $\boldsymbol{\pi}^*_0$ provides a basis for clinical decision support. Specifically, let $Y^{\pi}$ denote the counterfactual outcome under a hypothetical behavioral policy $\pi$, and let $Y^{j,\pi}$ denote the counterfactual outcome under a hypothetical provider assignment $j$ and hypothetical behavioral policy $\pi$.   The exclusion restriction on our causal model $\mathcal{M^F}$ implies that $J$ affects $Y$ only through $\boldsymbol{\pi}$, and thus that $\mathbb{E}_{O,U}[Y^j]\equiv \mathbb{E}_{O,U}[Y^{j,\boldsymbol{\pi}^j}]\equiv \mathbb{E}_{O,U}[Y^{\boldsymbol{\pi}^j}]$, and thus that
\begin{equation}
\mathbb{E}_{O,U}[Y^{d_j^*}]= \mathbb{E}_{O,U}[Y^{\boldsymbol{\pi}_0^*}],  
\end{equation}
or in other words, that the expected outcomes obtained were we able to assign each patient the most skilled provider, conditional on patient characteristics, can also be obtained by emulating the (observed) behavioral policy of these optimal providers. This is a particularly powerful result, in that it suggests a means of guiding multiple complex clinical decisions over time in response to a massive and evolving information set. In other words, rather than directly solving a complex sequential optimization problem (estimating a longitudinal optimal dynamic regime, see \cite{Moodie_2007} and \cite{Murphy_2003}), we instead leverage the fact that skilled clinicians are themselves implicitly solving such a problem based on extensive experience, and are revealing their solutions through their actions. 
\\
\\
While these results allow us to identify provider-specific policies and the causal effects of alternative policies on outcomes (given sufficient support), they do not directly identify the causal effects of specific actions (or of action regimes other than those followed by providers in the population). In other words, we learn who the highest quality providers are and how they would act given patient context, but not which specific actions or policy characteristics are important in driving quality. However, we can leverage the provider specific policies $\boldsymbol{\pi}^j_0$ and patient specific optimal policy $\boldsymbol{\pi}^{*}_0$ to establish a general framework in which we can identify the features of behavioral policies which drive the outcome distribution identified across providers. 

\section{Estimation}
We now outline our approach to estimating each of the estimands presented: the optimal provider assignment mechanism as a function of the provider information set; the value of (i.e., expected counterfactual outcome under) this provider assignment mechanism; the behavioral policy (i.e. a stochastic policy defined on the space of possible clinical actions); and, the (optimal) provider-specific behavioral policy. 

\subsection{Estimating the optimal provider assignment policy and its value}
In estimating the optimal provider assignment policy and the value of this policy, we borrow from existing literature on estimators for optimal individualized treatment rules (with the provider playing the role of ``treatment" (e.g.,\cite{Murphy_2003}, \cite{Moodie_2007}, \cite{Chakraborty_Moodie_DTR})). We briefly review several relevant estimation approaches here, noting that this is not an exhaustive list.  

\subsubsection{Single-stage Q-learning}
One simple approach is to construct a simple plug-in estimator of the  conditional average treatment effect (CATE, or ``blip function") generalized to multiple level categorical treatments.  Let  $\bar{Q}_0(J,I_J)$ denote $E_0(Y|J,I_J)$ and let
\begin{equation}
    B_0([j',j], I_J) \equiv \bar{Q}_0(j', I_J) - \bar{Q}_0(j, I_J),
\end{equation}
 denote the pairwise blip. This captures how much better, or worse, the expected outcomes of patients with information set $I_J$ assigned to physician $j'$ are compared to those assigned to physician $j$. We can construct the matrix of cross-provider outcomes for each pair of possible  assignments $j',j \in \mathcal{J} \times \mathcal{J}$.  Let $|\mathcal{J}| = m$, and define 
\begin{equation}
    \Tilde{B}_0(j', I_J) \equiv     
    \frac{1}{m}\sum_{j \in J}^{m} B_0([j',j], I_J) = \bar{Q}_0(j', I_J) - \frac{1}{m} \sum_{j \in J} \bar{Q}_0(j, I_J)
\end{equation}
Finally, let $\Tilde{B}_0(I_J) \equiv \{\Tilde{B}_0(j', I_J), j' \in \mathcal{J}\}$ denote a``pseudo-blip" (see \cite{Laan_Coyle_2023}), a vector of length $m$ that reflects how much better, or worse,  outcomes of patients with information set $I_J$ would be under assignment to  provider $j'$, compared to the average expected outcome under assignment to all providers. 
\\
\\
Maximizing $ \Tilde{B}_0(j', I_J)$ is equivalent to finding the provider assignment $j'$ such that $\bar{Q}_0(J = j', I_J)$ is maximized. 
\begin{equation}
    d^*_{J} = \text{argmax}_{d_J \in \mathcal{D}} \mathbb{E}_0 (\Tilde{B}_0(j, I_J))
\end{equation}
\textbf{Proof:} Notice that $\forall \; j^* \neq j'$: $\Tilde{B}_0(j^*, I_J) - \Tilde{B}_0(j', I_J) = \bar{Q}_0(j^*,I_J) - \bar{Q}_0(j',I_J)$ since the second term in the pseudo-blip is the same for all $j$. Then, by definition of the argmax, and assuming w.l.o.g that this is a singleton;
\begin{align*}
    d_J^*(I_J) & = \text{argmax}_{d_J \in \mathcal{D}} \mathbb{E}_0 (\Tilde{B}_0(d_J, I_J)) \\
        & \Rightarrow \mathbb{E}_0 \Big(\bar{Q}_0(d_J^*(I_J),I_J)\Big) > \mathbb{E}_0 \Big(\bar{Q}_0(j,I_J) \Big) \; \; \forall \; \; j \neq d_J^*(I_J)
\end{align*}
Then, by definition of $d^*_J$, since $\mathbb{E}_{0}(Y^{d_J}) = \mathbb{E}_0(\bar{Q}_0(J=d_J,I_J))$, the expression above follows $_\Box$. 
\\
\\
As such, given an estimator for $\Tilde{B}_n(j', I_J)$, we can find the optimal provider assignment policy. Note that this extends to any such optimization process using categorical treatment and the relevant pseudo-blip estimator. One simple option is to estimate $\Tilde{B}_0(j', I_J)$ using a simple plug-in estimator $\bar{Q}_n(J,I_J)$ of $\bar{Q}_0(J,I_J)$. However, a limitation of this approach is that performance depends entirely on the performance of the estimator of $\bar{Q}_n(J,I_J)$.

\subsubsection{Direct estimation of the value of candidate provider assignment policies}
An alternative approach is to directly estimate the value of candidate provider assignment policies $d_j \in \cal{D}$. Here, double robust (or semiparametric efficient) approaches are particularly appealing due to their ability to incorporate machine learning-based estimators, and in particular neural network-based approaches, to capture the rich multimodal data in estimating both the provider assignment mechanism $g_0$ and the conditional expectation of the outcome $\bar{Q}_0$ while maintaining desirable asymptotic properties. Throughout, we assume appropriate internal sample splitting (cross-validation or cross-fitting) is employed; for readability, we omit full details and notation of these procedures and point readers to \cite{AtheyWagner2021}, \cite{Zheng_2011}, \cite{Luedtke_Laan_2016}. 
\\
\\
One double robust approach to directly estimating the value of a candidate policy is the Augmented Inverse Probability Weighted (A-IPW) estimator (see \cite{Bang_2005}). Given estimators $g_n$ of $g_0$ and $\bar{Q}_n$ of $\bar{Q}_0$, the expected patient outcome under a candidate provider assignment mechanism $d_J(I_J)$ can be estimated as;  
\begin{equation*}
    \mathbb{E}_n[Y^{d_J}]=\frac{1}{n}\sum_{i=1}^n \Bigg(\frac{\mathbb{I}(J_i = d_J(I_{J,i}))}{g_{0}(J_i|I_{J,i})} (Y_i - \bar{Q}_{0}(J_i,I_{J,i})) + \bar{Q}_{0}(d_J(I_{J,i}), I_{J,i})\Bigg) 
\end{equation*}
Alternatively, one can consider a Targeted Maximum Likelihood Estimator (TMLE) of the value of a candidate rule (e.g. \cite{Laan_Rose_TargetedLearning}).  In our setting the ``clever covariate" (or weight) of the TMLE is given by; 
\begin{equation}
    H_{n,i} = \frac{\mathbb{I}(J_i = d_J(I_{J,i}))}{g_n(J_i|I_{J,i})},
\end{equation}
After the appropriate logit transformation this leads to a targeted estimate of the conditional expected outcome $Q^*_n(d_J(I)|I)$. This is used to construct a plug-in estimator of the expected outcome under the candidate provider assignment rule: 
\begin{equation}
    \mathbb{E}_n(Y^{d_J}) = \frac{1}{n} \sum_{i=1}^{n} Q^*_n(d(I_{J,i}),I_{J,i}) 
\end{equation}
Either double robust estimator of the value of a candidate policy can be used to directly search the candidate policy space for the provider assignment policy that results in the highest estimated value:
\begin{equation}
    d^*_{J,n}(I_J) =\text{argmax}_{d_J \in \mathcal{D}} \mathbb{E}_n(Y^{d_J})
\end{equation}
Each of these approaches can further be used to estimate the value of the (learned) optimal provider assignment policy $E_0(Y^{d_{j,n}^*})$, or under additional conditions, the value of the true optimal policy $E_0(Y^{d_{j}^*})$ (again, assuming appropriate sample splitting as in \cite{Luedtke_Laan_TargetedMeanOutcomeODTR}). It remains, however, to define machine-learning approaches to searching the candidate policy space. A wide range of such approaches are available. 

\subsubsection{Super-learners of the optimal provider assignment policy}
A Super Learner approach can be used to effectively leverage the wide range of candidate estimators of the optimal provider assignment policy that are currently available: (see \cite{Luedtke_Laan_SuperLearning_ODTR}, \cite{Montoya_2022}, \cite{Laan_2003_UnifiedCM}). At its core, Super learning uses cross-validation to select among  candidate estimators of the optimal rule, as well as among combinations of these estimators. Specification of a Super Learning estimator of the optimal provider assignment policy  requires specification of a library of candidate estimators, a meta-learning approach for combining these estimators, and choice of a loss function. Given the high-dimensional categorical space of the candidate provider set considered, one particularly applicable Super Learner would consider a library of candidate estimators of the pseudo-blip, a (pseudo-blip)-based metalearner, and a squared error loss function. 

\subsection{Estimating the behavioral policy}
In this section we consider an approach to estimating the behavioral policy $\boldsymbol{\pi}_0$ and $\boldsymbol{\pi}_0^j$. Recall that $\boldsymbol{\pi}_0$ is defined as the probability distribution over the space of potential clinical actions for a given information-set, and as such defined as a product over period-$t$ specific policies. We  present an estimation strategy in which we do not need to estimate each of these objects separately. 
\\
\\
Estimation of $\boldsymbol{\pi}_0$ corresponds to a prediction problem in which an estimator is trained to predict over the space of period $t$ clinical actions, i.e. $A_{t}$, given the history of past clinical actions $\boldsymbol{A}_{0:t-1}$, patient states $\boldsymbol{X}_{0:t}$ and provider history $\boldsymbol{J}_{0:K}$; in other words, the model learns to fit the information-set conditional distribution of period-$t$ clinical actions, $\pi_t(A_t \; | \; \boldsymbol{I}_{0:t})$. Recent advances in the AI literature, especially NLP, have demonstrated the ability of large transformer-based models to learn rich representations of sequence data across multiple domains (\cite{TransformerAppSurvey}). Of particular interest is work on similar EHR token sequences, which has demonstrated the ability for large neural network architectures to extract useful patient representations (\cite{EHR_BERT}, \cite{MedBERT}, \cite{MOTOR}, \cite{ETHOS}, \cite{EHRMamba}, \cite{CLIMBR}). In our setting, we have three types of potentially multimodal and high-dimensional inputs which form the information set, a time-series of patient states\footnote{Note that we expand on this definition below as one needs to consider efficient embedding strategies to share this data with a given model.} $\boldsymbol{X}_{0:t}$, a time-series of previous clinical actions $\boldsymbol{A}_{0:t}$ and a time series of providers $\boldsymbol{J}_{0:K}$. The temporal nature of our inputs, in which information sets grow over time as more patient states and actions realize, motivates a sequence-to-sequence architecture (\cite{SeqToSeq}, \cite{MachineTranslation}). That is, we require an estimator that can learn the mapping from recursively updated information sets to future actions. As we will outline below, a transformer architecture, though by no means the only available estimator, presents a good candidate for fitting this complex clinical action mechanism. 
\\
\\
We separate the task of estimating $\boldsymbol{\pi}_{0}$ and $\boldsymbol{\pi}_{0}^j$ into two phases.  First, in an initial pre-training phase, we estimate a modified behavioral policy (or action mechanism) that differs from $\boldsymbol{\pi}_{0}$ by 1) excluding provider history from the inputs, and 2) spanning the full available action and state-space history, rather than indexing on a single encounter date $K$. Denote the corresponding pre-trained ``general" behavioral policy $\boldsymbol{\pi}_0^{\text{pre}} \equiv \prod_{t=0}^T\pi_t'(A_t \mid \boldsymbol{X}_{0:t},\boldsymbol{A}_{0:t-1})$ and an estimator of this policy $\boldsymbol{\pi}_n^{\text{pre}}$. The advantage of this approach is that it allows for a flexible underlying large behavioral model, for which predictions can be applied without access to underlying provider history as an input (An analogous argument can be made for defining the information set $I_J$ used for optimal provider assignment to exclude past provider history). This pre-trained model can be reused through varied fine tuning and sub-setting on encounter types for a variety of uses without re-training (\cite{PreTrainingSurvey}). In a second stage, the estimator of this general pre-trained policy is fine-tuned to a provider-specific behavioral policy.

\subsubsection{Transformer architecture}
Transformer architectures are a particular class of neural-network based estimators that can be used to estimate $\boldsymbol{\pi}_{0}$. This neural network architecture naturally handles sequential inputs and has found wide applicability in NLP and other sequence-to-sequence tasks (\cite{AttentionIsAl}, \cite{BERT}, \cite{EHR_BERT}, \cite{MedBERT}, \cite{GPT}, \cite{ViT_VisionTransformer}, \cite{SurveyOfTransformers}). In this section, we  provide a definition of the standard attention-based transformer architecture, as in \cite{AttentionIsAl}, as a sequence of transformer blocks, noting that this is by no means the only transformer-style architecture, nor is it the only sequence model applicable to our data setting. 
\\
\\
A transformer consists of a series of transformer blocks. Each transformer block is a function which maps a sequence of inputs to a sequence of outputs. In our setting, this is a sequence of clinical actions and patient states being mapped to future clinical actions, where we can leverage a large literature on multimodal transformer architectures, see \cite{MultimodalTransformers}, to capture $\boldsymbol{A}_{0:t}$,  $\boldsymbol{X}_{0:t}$.   
\\
\\
To prepare the input data, consider an embedding layer $e$ which maps the $t$-period information set $\boldsymbol{I}_{0:t}$ to a $d \times T$ dimensional embedding sequence, i.e. $e(\boldsymbol{I}_{0:t}) \in \mathbb{R}^{d\times T}$, where $T - t$ is added as padding. We propose a general embedding approach in the section below. Additionally, to allow the network to leverage the positional information in the input sequence, i.e. which actions are preceding others, a positional-encoding $\boldsymbol{P} \in \mathbb{R}^{n \times T}$ are added to the input embeddings (see \cite{AttentionIsAl} and \cite{PosEncoding} for a treatment of standard positional encoding and \cite{RoFormer} for a rotation-based approach). The transformer block then consists of two layers, a self-attention layer and a point-wise feed forward layer. The attention layer computes a mapping between pieces of the input embedding sequence as they relate to the prediction task at hand. A complete transformer block is then a function $f_{\theta}: \mathbb{R}^{d \times T} \rightarrow \mathbb{R}^{d \times T}$ defined by the hyperparameters $\theta \equiv (h, m, d, r)$. These are the number of heads in the self-attention layer $h$, the size of each head $m$, as well as the embedding dimension $d$ and the hidden layer dimension $r$ of the feed-forward layer. We demonstrate that even a simple model of this kind is well-suited for our data setting in Section 4 below. 
\\
\\
Connecting this back to our causal model, the attention layers in the transformer architecture capture the arbitrarily complex causal (and by implication, statistical) relationships between $\boldsymbol{X}_{0:t-1}, \boldsymbol{A}_{0:t-1}, \boldsymbol{J}_{0:K}$ and $A_{t}$. Having a highly data-adaptive class of estimators that can easily handle complex time-varying relationships between high-dimensional features is at the core of what makes our approach empirically feasible. although there are other estimators, particularly other neutral network architectures, which can plausibly be applied to our setting as well, the transformer architecture, with attention at its core, has shown to be highly capable across a range of sequence-to-sequence domains and presents a natural fit for our estimation problem.

\subsubsection{Training}
We now consider how to estimate a transformer model of the kind outlined above. We begin by setting up the pre-training task. Here, one needs to compile a set of training data which consist of information set and next action pairs, that is, define a dataset: $\boldsymbol{D} \equiv \{I_{i,0:t}, A_{i, t+1}\}_{i,t}^{N,T}$, where $N \times (T+1)$ is the number of such information-set and action pairs used to fit the model. During pre-training the information set does not include provider-specific indicators because we are sampling action paths from the entire population. The model is then scored on its predictions of the next set of clinical actions, and a candidate generative model $\boldsymbol{\pi}_n^{\text{pre}}$ learns the combination of weights, i.e. parameters, which minimize the difference between predicted and actual sequences of actions. This ``next-action" prediction task reliably embeds the desired sequence-to-sequence behavior in this class of model, mirroring the way in which large language models are trained by predicting the next word in a sentence (\cite{GPT}, \cite{GPT4_Report}). Our general behavioral policy estimator $\boldsymbol{\pi}_n^{\text{pre}}$ is learning to represent the clinical action mechanism, that is, the sequence of actions likely to be taken by physicians with access to patient-specific information set $I_{i, t}$.
\\
\\
After estimating $\boldsymbol{\pi}_0^{\text{pre}}$ without including provider history, we now return to the task of fine-tuning\footnote{Note that we are using this to mean any method by which one aligns a pre-trained model to provider-specific behavioral patterns. Other methods, especially reinforcement learning techniques, are also available;  we focus here on fine-tuning for its expositional simplicity.} to the policies of specific providers, yielding estimates for the action path distribution conditional on assignment to a given provider, denoted by $\boldsymbol{\pi}_0^j$. The key here, as before, is to construct a dataset that now consists of the information sets and actions taken by a given provider $j$, i.e. $\boldsymbol{D}_j \equiv \{I_{i,0:t}, A_{i,t+1} \; | \; J = j\}_{i,t}^{N_j,T}$. Note that $N_j$ is the number of patients which were assigned to $j$, and that there should be previously un-used data in $\boldsymbol{D}_j$, i.e. encounters which were not used during pre-training, such that new signal is available for the model during this phase of training. The training task remains the same as the model learns to predict the next set of clinical actions along a path of real actions taken by a provider $j$. In Section 4 we demonstrate the ability of a standard transformer architecture to learn to predict the distribution of likely next clinical actions. 

\subsubsection{General multimodal embedding}
One reason why transformer architectures are a promising approach is their ability to take in rich multimodal inputs (\cite{MultimodalTransformers}). Though the vast majority of existing research focuses on combining modalities such as images and text, there nothing fundamentally different about the multimodality of patient states and actions. Generally, one can view these kinds of problems as the model-sourcing inputs from separate data manifolds, i.e. the space of images vs. the space of text, which is equally the case for provider actions and patient states. Although the exact implementation details are beyond the scope of this paper, we briefly introduce a general embedding approach to dealing with the kinds of multimodal time-series data we consider in our research. 
\\
\\
In direct correspondence with the way clinical data are generated, we let $X(A) \in \mathcal{X}$ be the state of action $A$ associated with a patient. Here we expand on the definition of patient states used above. This approach is motivated by its information efficiency; because multiple actions can share the same state,  the size of a given model's vocabulary (i.e., size of the required action space) is reduced and the efficiency with which multimodal data is ingested is increased. A key observation here is that a patient's state is strictly defined, and observed, through an action. That is, a change in a patient's blood pressure, i.e., change in patient state, is observed only because a provider took the patient's blood pressure, i.e. choose an action to perform at a point in time. A simple example is to consider lab tests. These are actions that return a result, that is, a lab text action $A$ can be performed across multiple patients and return different results, i.e. $X_{i}(A) \in \text{``Possible Results"}$, for patients indexed by $i$. In this way, we are able to capture essentially all information contained in patient order paths in a succinct state-space representation in which each action maps neatly to an associated state. Different actions can map to different state-spaces, i.e. lab results vs. vital signs vs. medication dosages, and these can be differentiated along both "type" as well as whether or not a given state is chosen or realized exogenously (i.e. lab results). Note that the null-state is a valid state in $\mathcal{X}$, so that actions without direct state-space interpretations fit this setting as well. Some other examples of natural state-space representations include recording diagnosis (i.e. $A_{t}$ is the act of recording a differential and $X(A_{t})$ maps to the relevant ICD-10 code recorded), prescribing medication (i.e. $A_{t}$ is the medication being prescribed and $X(A_{t})$ is the duration, dosage, and intervals of the prescription), and changes in treatment location (i.e. $A_{t}$ is moving the patient and $X(A_{t})$ maps to ER, in-patient, ICU, etc.). A given information set $\boldsymbol{I}_{0:t}$ is then embedded in two separate spaces giving rise to vectors over $\mathcal{A}$ and $\mathcal{X}$ (or versions thereof), which can be combined using any number of ways, including concatenation, cross-attention, or simple summation to form our final embedding layer $e(I)$. Though we do not make use of this strategy in this paper, since we present a simple unimodal model, this is an ongoing area of research. 

\subsubsection{Universal approximation properties}
We now consider an additional feature of transformer architectures which motivate their use as an estimator of $\boldsymbol{\pi}_0$. As established in \cite{Yun2020}, transformers are universal approximators of sequence-to-sequence functions, which is exactly what we require for an estimator of the complex clinical action mechanism. Specifically, under conditions outlined in \cite{Yun2020} it is possible to show that any function in the set of continuous and compact functions from $\mathbb{R}^{n \times T} \rightarrow \mathbb{R}^{n \times T}$ can be arbitrarily closely approximated by a transformer architecture with a positional encoding layer and the corresponding embedding dimension $n$ and context window $T$. With enough data and sufficient parameter tuning the above result implies that a transformer architecture is well suited to estimate the clinical action mechanism we outline in section 2. That is, with enough data we are able to estimate a transformer architecture $\boldsymbol{\pi}_n$ such that the clinical action mechanism implied by it and the data we observe is arbitrarily small. Note here that, as outlined in the remark above, we do not need to separately estimate period-specific mechanisms. 

\begin{remark}(\textit{Sequence Decoding})
    Note that, as this lies outside the scope of this paper, we are abstracting away from the process by which a sequence-to-sequence transformer model can be used to predict sequences of individual actions, i.e. $A_{t:t+z}$ from $I_{0:t}$. For the purposes of the theory and approach presented in this paper it suffices to establish that such a model is capable of learning $\pi_0(A_t \; | \; I_{0:t})$ from complex multimodal data, while noting that sequences of actions can be predicted by sampling from these learned distributions recursively (see \cite{SeqToSeq} for a general treatment). 
\end{remark}

\subsubsection{Deep Causal Behavioral Policy Learning (DC-BPL)}
We now summarize the above pieces into our \textbf{deep causal behavioral policy learning} algorithm. Recall, as established in section 2, that under the assumption of conditional exchangeability of $J$ we can identify the expected counterfactual outcome under assignment to some $j \in \mathcal{J}$, the optimal provider $d^*_J(I_J)$ as a function of $I_J$ (the information set at the time the provider is assigned), and the optimal provider's counterfactual behavioral policy $\boldsymbol{\pi}^*_{0}$. 
\begin{algorithm}[H]
\caption{Deep Causal Behavioral Policy Learning (DC-BPL)}\label{alg:lbm_alg}
\begin{algorithmic}
\Require Estimators for $d^*_{J}$ and $\boldsymbol{\pi}_n$, $\mathcal{J}$, $\boldsymbol{D} \equiv \{I_{i,0:t}, A_{i, t+1}\}_{i,t = 0}^{N,T}$, $\boldsymbol{D}_j \equiv \{I_{i,0:t}, A_{i,t+1} \; | \; J = j\}_{i,t}^{N_j,T}$
    \begin{enumerate}
        \item Fit baseline transformer $\boldsymbol{\pi}_n^{\text{pre}}$ on $\boldsymbol{D}$.
        \item Separately fine-tune $\boldsymbol{\pi}_n^{\text{pre}}$ on $\boldsymbol{D}_j$ in order to construct a series of estimators $\boldsymbol{\pi}_{n}^{j}$, which are the $j$-specific behavioral policies. 
        \item Estimate $d^*_{J,n}(I)$ for $I \in \mathcal{I}$.
        \item The optimal causal BP-estimator for $I$ is then; $\boldsymbol{\pi}^*_n \equiv \boldsymbol{\pi}_n^{d^*_{J,n}(I)}$
    \end{enumerate}
\end{algorithmic}
\end{algorithm}

\subsubsection{Consistency of DC-BPL}
The assumptions required for consistency (and other statistical properties) of the estimators presented in Section 3.1 for the optimal   provider assignment rule for a given patient information set are well-studied. What is less established is the ability to consistently fit an estimator of the behavioral policy $\boldsymbol{\pi}_n^{\text{pre}}$ and individual behavioral policy $\boldsymbol{\pi}_n^j$. To our knowledge there is no established asymptotic theory for transformer style networks as statistical estimators of this kind, though, as mentioned above, we can rely on the existence of the ``correct" transformer-based network (\cite{Yun2020}). The closest existing results are the widely publicized scaling laws of transformers applied to text data (another complex sequence domain), in which more data leads to better out -of-sample fit, but without guarantees for asymptotic consistency (\cite{ScalingLaws} introduced scaling laws in this domain and recently \cite{ScalingLawStatistics} has given this a rigorous statistical treatment). In Section 4 we present empirical analyses investigating the performance of a transformer architecture to estimate $\boldsymbol{\pi}_0$. 

\section{Empirical Analysis: Estimating the LCBM}
In this section we present results from a proof-of-concept analysis applying a simple transformer to estimate a behavioral policy using a sample of electronic health record data from a tertiary care Emergency Department. Our analysis uses a simpler unimodal architecture (compared to Section 3.2 and without multimodal embedding as in 3.2.3) in order  to establish a baseline of performance. 
Prior related analyses, with the partial exception of \cite{EHRMamba}, have primarily focused on leveraging these order sequences to learn representations of patients for use in downstream prediction tasks (\cite{CLIMBR}, \cite{ETHOS}, \cite{EHRMamba}, \cite{MOTOR}). In contrast, we evalaute the models' ability to learn the actual underlying clinical action mechanism, which we believe embeds rich clinical logic.  We propose and implement several novel ways to evaluate model performance for this task. 

\subsection{Model Architecture}
As a demonstration of the ability of transformer architectures to learn the clinical action mechanism, we set up a unimodal pre-train task for a basic sequence-2-sequence model. We employ an encoder-decoder architecture to learn the mapping from previous actions along a patient's path to the next \textit{set} of actions. An important feature of our data, which can be integrated into the models above, is that in many settings multiple actions are recorded at the same time and/or within short horizons. For example, it is often the case that multiple labs are placed at the same time. In this example implementation, our model learns to predict an order-set $A_{t+1:t+z}$ recursively from input time-series $A_{1:t}$ where $z$ is the size of the target order-set. Note that this differs from the set-up in \cite{EHRMamba}, \cite{ETHOS}, and \cite{MOTOR}, but is markedly closer to the process with which clinical decisions are made. 
\\
\\
Treating the order-path data as a series of sets adds significant complexity; we explore simultaneous multi-label style predictions over next actions in ongoing work. The model's context window defines the maximum $t$ and $z$,  i.e., the patient history of actions and size of future order sets. We employ a standard tokenizer and learned embedding layer to map individual actions in $\mathcal{A}$, which forms our vocabulary, to embedding vectors in $\mathbb{R}^{d}$. We add standard positional encoding and use standard causal attention in which the model is recursively predicting actions in the target order-set $A_{t+1:t+z}$ with full access to all previous action embeddings. For the sake of simplicity, we use a uni-modal model here, that is, we do not make use of any states $X$. 

\subsection{Sample training task}
We train our model on a dataset of encounter-level action path sequences, which are structured as pairs of sets. For each encounter, we generate recursively increasing input sets to capture the set of actions that have taken place leading up to a particular point in time. We then form batches from these action-sequence pairs and ensure that our train/test split occurs at the encounter-level, such that there is no leakage of action-sequence pairs across datasets. Using the notation above, this leaves us with a pre-training dataset defined as $\boldsymbol{D} = \{ A_{i, 0:t}, A_{i, t+1:z} \} \; \text{ across $i$ and $t$}$. We only consider the pre-training task here; a full implementation of our causal fine-tuning algorithm lies outside the scope of this paper. As such, there is no provider-specific fine tuning data and we evaluate our model entirely on its ability to learn the ``general" (provider-agnostic) action mechanism. We present results from applying this architecture to a training set of 180,000 unique UCSF Emergency Department encounters covering 115,000 patients with one of the following common chief-complaints: \textit{Abdominal pain, Fever, Cough, Emesis, Chest pain}, and \textit{Shortness of breath}. Our action space includes the 900 most common procedural actions in the data and we do no other data pre-processing. This data is provided as part of the UCSF Information Commons (\cite{IC_CDW}). 
\\
\\
It is likely that performance could be markedly improved with additional data pre-processing, more data, and larger vocabularies. This leaves us with $1,668,872$ unique order set pairs to train on, with an average number of $88.9$ input tokens and $3.0$ targets. We train a small 53 million parameter model on these unimodal data with $d = 1024$, $h =4$, $r = d = 1024$, $3$ decoder and encoder layers, a batch-size of $16$, a maximum window size of $512$, and a small constant learning rate at $1.0e-8$. This is as simple a set up as possible;  other architectures are explored in existing literature (although with different objectives, such as in \cite{EHRMamba} and \cite{MOTOR}). The objective of this proof-of-concept analysis is to demonstrate that a fairly small and simple model of this kind, without any additional data processing or cleaning, is able to capture significant signal from the sequential clinical decision process. 

\subsection{Initial model evaluation}
In this section we present an evaluation of the above model. We do not impose a specific class of decoder, but instead evaluate the performance of our estimator $\boldsymbol{\pi}_n$ directly, i.e. we evaluate the ``raw" predicted probability distribution since this is the core object of interest for a stochastic longitudinal behavioral policy. For this reason, we do not make use of the standard multi-label accuracy measures (i.e. precision, recall, F$_1$, etc.) and instead rely on metrics which can be defined on the predicted distribution over $\mathcal{A}$ directly. For notational convenience we will denote the target set of actions by $A'$. We will consider different features of the prediction setting to evaluate how performance metrics behave as a function of (1) a feature we call ``learned separation" and (2) context length $t$ . We define two metrics for prediction performance: (1) mean and min-top-k accuracy; and, (2) the quantile function for actions in $A'$. 

\subsubsection{Action-level learned separation}
In this section we measure how well $\boldsymbol{\pi}^{\text{pre}}_n$ has learned to separate when a particular action should and should not be predicted in the next set. We analyze the predicted probabilities of actions when a given action $a$ is selected by a provider and when it is not. Here we sample 300,000 order-set pairs not used during training and evaluate our BPL-estimator by considering moments of the empirical CDF of probabilities assigned to actions when they are vs. when they are not placed. We denote these as: 
\begin{align}
    {F}_n^{(1)}(a) &\equiv F_n(\boldsymbol{\pi}^{\text{pre}}_n(I)[a] \; | \; a \in A') \\
    {F}_n^{(0)}(a) &\equiv F_n(\boldsymbol{\pi}^{\text{pre}}_n(I)[a] \; | \; a \notin A')
\end{align}
Here, $\boldsymbol{\pi}^{\text{pre}}_n(I)[a]$ is the predicted probability of action $a$ under information set (i.e. previous action path) $I$. We denote CDFs by $F$ and distributions across all actions  $a \in \mathcal{A}$ as ${F}_n^{(1)}$ and ${F}_n^{(0)}$ respectively. For a functioning decoder we require some sufficient degree of separation between these two distributions, i.e. the model needs to have learned to assign different probabilities to the same action when it does and doesn't occur in $A'$. 
\\
\\
When applied to our proof-of-concept pre-trained model we find that the mean value of ${F}_n^{(0)}$ is $0.0005$ and the mean value of ${F}_n^{(1)}$ is $0.005$. This result means that an action $a$, on average, receives an approximately 10-times higher predicted probability when it does occur in the next order set (i.e. the distribution for $a \in A'$), than when it does not. The key to this metric is that we can break this analysis down to the action-level. To our knowledge, this is the first use of the action-level ``learned separation" in this way, as we establish a simple difference-in-means statistic which, at the action-by-action level, is highly predictive of the model's accuracy (other moments of the action-level CDFs are also possible and themselves interesting.)  This approach is based on the fact that the model will have been exposed to all actions in the vocabulary at different rates and across different information-sets. It stands to reason that some actions are ``easier" to learn than others, and that actions which are easier to learn are those in which the model can more easily distinguish between when they should and should not be predicted next. In this sense, "learned separation" provides a measure of the model's degree of certainty when making a prediction over a given action $a$.  
\\
\\
Let $\boldsymbol{D}_{\text{eval}, \;a}^{(1)}$ be a data-set of order-paths in which some action $a$ occurs in the next order set (i.e. should be predicted to occur) and let $\boldsymbol{D}_{\text{eval}, \;a}^{(0)}$ be a data-set of order-paths in which $a$ does not occur next. These are both calibration data-sets not used during training on which we compute the following difference in means statistic; 
\begin{equation}
    \Delta^{\pi_n}(a) \equiv \frac{1}{|\boldsymbol{D}_{\text{eval},\;a}^{(1)}|} \sum_{I_i \in \boldsymbol{D}_{\text{eval},a}^{(1)}} \boldsymbol{\pi}^{\text{pre}}_n(I_i)[a] - \frac{1}{|\boldsymbol{D}_{\text{eval},\;a}^{(0)}|} \sum_{I_j \in \boldsymbol{D}_{\text{eval},a}^{(0)}} \boldsymbol{\pi}^{\text{pre}}_n(I_j)[a] 
\end{equation}
Note that $|\boldsymbol{D}_{\text{eval},a}^{(1)}| < |\boldsymbol{D}_{\text{eval},a}^{(0)}|$, since for most actions there are many more order sets in which they don't occur, and we sample sets of the same size for each. 
\\
\\
When applied to our fitted model we find that \textbf{79.5\%} of our action space the model has learned positive separation between mean predicted probabilities, i.e. $\Delta^{\pi_n}(a) > 0$; of these differences, 79.3\% are significant at the 5\% level using a difference in means test. Although we omit the plot here for brevity, the learned mean separation displays a log-log linear relationship with the relative frequency with which a given action is observed during training. In other words, learned separation and order frequency are positively correlated. As we demonstrate below, however, $\Delta^\pi$ is more predictive of model performance than frequency itself, since the proposed learned separation metrics  accounts for how ``well" the model has learned to differentiate for a given action. 
\begin{remark}
We propose using this approach to markedly reduce, and eventually eliminating, hallucinations for models deployed in expert domains such as clinical decision support. With as simple a feature as $\Delta^\pi$ we are able to limit our model's responses to the subset of the action space in which we are confident that accuracy is high. As we demonstrate below, $\Delta^\pi$ is highly predictive of model accuracy, and allows us to identify settings where model performance is low/high.
\end{remark}

\subsubsection{Mean/Min Top-k Accuracy}
In this section we consider mean and min top-k accuracy across a sample of 300,000 action set pairs not used during training. The mean top-k accuracy is computed by considering the mean rank of $a \in A'$, that is the mean placement of actions in the target set under the predicted distribution, and comparing against an integer $k$. As such we compute, across an evaluation dataset, the following statistic: 
\begin{equation}
    \text{mean top-k} \equiv \frac{1}{|\boldsymbol{D}_{\text{eval}}|} \sum_{A' \in \boldsymbol{D}} \mathbb{I}\Big\{ \Big(\frac{1}{|A'|} \sum_{a \in A'} \text{loc}[a\;|\;I] \Big) \leq k \Big\}
\end{equation}
Here $\boldsymbol{D}_{\text{eval}}$ is a dataset of action-path pairs and $\text{loc}[\dots]$ is a function which takes the location of action $a$ in the predicted probability distribution of $\boldsymbol{\pi}^{\text{pre}}_n$ conditional on $I$. As such $\text{loc}[a] = 0$ implies that the action $a$ received the highest probability in $\boldsymbol{\pi}_n(I)$. In other words, $\text{loc}[a|I]$ is the location of $a$ in the order-statistic on $\mathcal{A}$ implied by $\boldsymbol{\pi}^{\text{pre}}_n(A|I)$ which is the learned distribution of our BPL-estimator at a given input where $I = \boldsymbol{A}_{0:t}$ in this simple set-up. As such, for $k = 10$, this computes the probability that the true actions lie within the top-10 highest predicted next actions over set of test data $\boldsymbol{D}_{\text{eval}}$. Below we plot the full degree of variation across $t \leq 512$ as well as the mean for $t \geq 200$. Note that increased variation seen with higher values of $t$ is due to decreasing data support. 
\begin{center}
    \includegraphics[scale = 0.5]{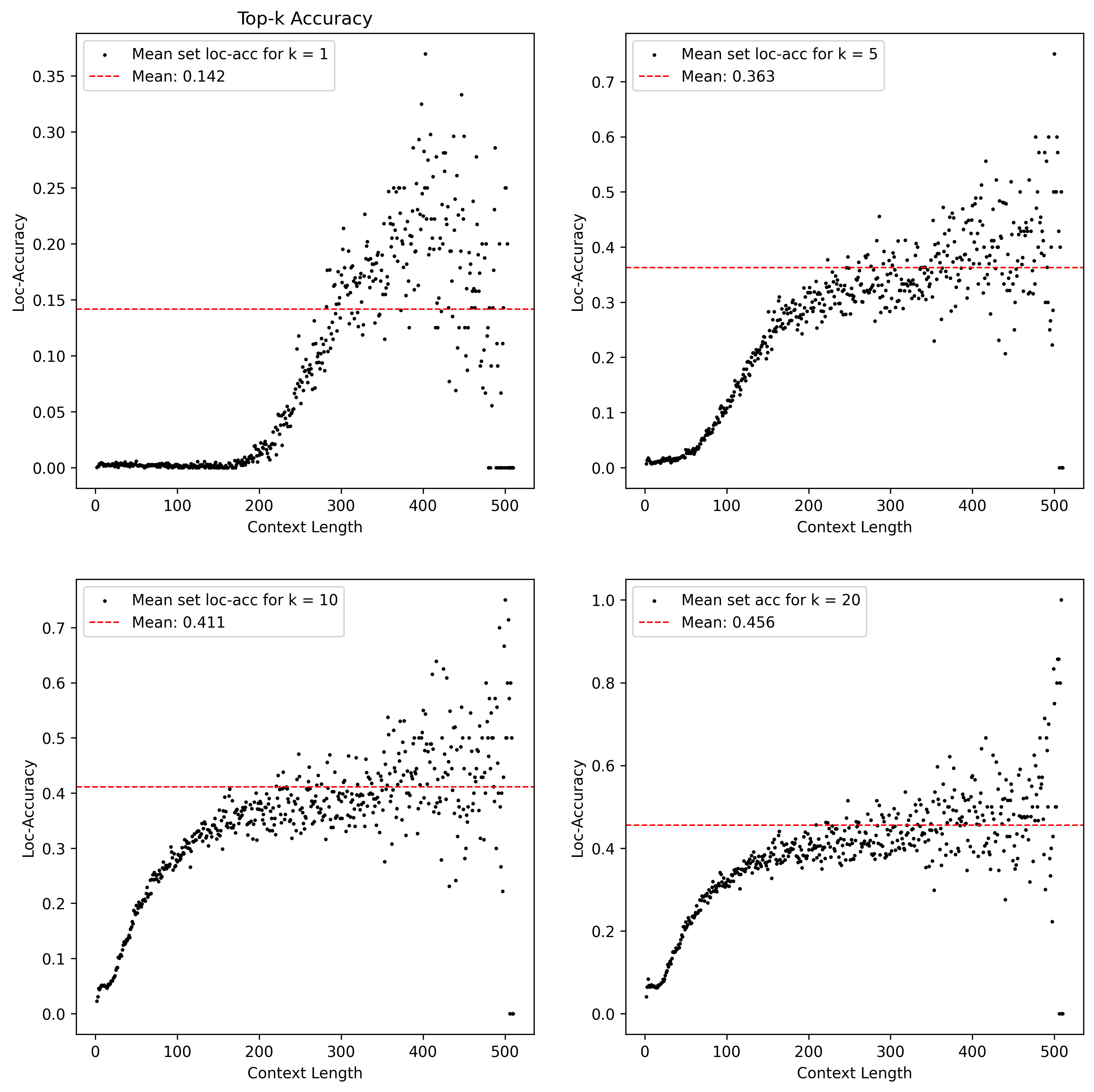}\\
    \textbf{Figure 1:} Top-k mean accuracy by $t$
\end{center}
As is clear from the figure above (and the table 1 below), more context, i.e. higher $t$ at the point of prediction, leads to markedly higher predictive accuracy. This speaks to, among other things, the models ability to extract useful information, i.e. signal over noise, from increasingly complex context.
\\
\\
In addition to the mean top-k accuracy we also compute the min-top-k accuracy and report differences below. The min top-k accuracy replaces the mean location inside the indicator in (20) with the minimum location over $a \in A'$, i.e. we compute; 
\begin{equation}
    \text{min top-k} \equiv \frac{1}{|\boldsymbol{D}_{\text{eval}}|} \sum_{A' \in \boldsymbol{D}} \mathbb{I}\Big\{ \text{min}_{a \in A'} \; \text{loc}[a\;|\;I] \Big) \leq k \Big\}
\end{equation}
Below we present tables that summarize the top-$k$ performance for both mean and min for varying context lengths. These results indicate significant increases in the model's ability to predict actions in $A'$ when $t$ increases. Moreover, the min accuracy is markedly higher than the mean accuracy (especially for small $t$) which points to the fact that $\boldsymbol{\pi}^{\text{pre}}_n$ is able to correctly predict at least one action in $A'$ significantly more often than all actions in $A'$ (i.e. the ``mean" action in $A'$).  
\begin{center}
    \begin{table}[h]
    \centering
        \vspace{-5pt} 
        \caption{Mean-top-$k$ Accuracy for next-action prediction} 
        \begin{tabular}{l l l l l l } 
             \hline
             Context Length &  \multicolumn{5}{c}{Accuracy}  \\
              & k = 1 & k = 5 & k = 10 &  k = 15 & k = 20  \\ [0.5ex] 
             \hline\hline
             $ t < 50 $             & 0.0026 & 0.0139 & 0.0926 & 0.0972 & 0.1116  \\ 	
             $ 50 \geq t < 100 $    & 0.0020 & 0.0599 & 0.2416 & 0.2492 & 0.2756  \\ 	
             $ 100 \geq t < 150 $   & 0.0016 & 0.1746 & 0.3158 & 0.3326 & 0.3518  \\ 	
             $ 150 \geq t < 200 $   & 0.0048 & 0.2726 & 0.3565 & 0.3752 & 0.3910  \\   
             $ 200 \geq t < 250 $   & 0.0377 & 0.3135 & 0.3764 & 0.3965 & 0.4121  \\	
             $ 250 \geq t < 300 $   & 0.1093 & 0.3327 & 0.3803 & 0.4082 & 0.4224  \\
             $ 300 \geq t < 350 $   & 0.1676 & 0.3423 & 0.3908 & 0.4202 & 0.4404  \\   
             $ 350 \geq t < 400 $   & \textbf{0.2061} & 0.3792 & 0.4284 & 0.4539 & 0.4700  \\   
             $ 400 \geq t $         & 0.1606 & \textbf{0.4000} & \textbf{0.4409} & \textbf{0.4745} & \textbf{0.4910}  \\   
        \end{tabular}
        \vspace{-25pt} 
    \end{table}
\end{center}
\begin{center}
    \begin{table}[h]
    \centering
        \vspace{-25pt} 
        \caption{Min-top-$k$ Accuracy for next-action prediction} 
        \begin{tabular}{l l l l l l } 
             \hline
             Context Length &  \multicolumn{5}{c}{Accuracy}  \\
              & k = 1 & k = 5 & k = 10 &  k = 15 & k = 20  \\ [0.5ex] 
             \hline\hline
             $ t < 50 $             & 0.0367 & 0.0636 & 0.2123 & 0.2208 & 0.2404 \\ 	
             $ 50 \geq t < 100 $    & 0.0239 & 0.0979 & 0.3179 & 0.3342 & 0.3715 \\ 	
             $ 100 \geq t < 150 $   & 0.0231 & 0.2129 & 0.3761 & 0.4180 & 0.4484 \\ 	
             $ 150 \geq t < 200 $   & 0.0285 & 0.3098 & 0.4089 & 0.4535 & 0.4851 \\   
             $ 200 \geq t < 250 $   & 0.0623 & 0.3550 & 0.4374 & 0.4805 & 0.5153 \\	
             $ 250 \geq t < 300 $   & 0.1350 & 0.3787 & 0.4482 & 0.5074 & 0.5466 \\
             $ 300 \geq t < 350 $   & 0.1954 & 0.4044 & 0.4858 & 0.5469 & 0.5771 \\   
             $ 350 \geq t < 400 $   & \textbf{0.2397} & 0.4563 & 0.5334 & 0.5795 & 0.6040 \\
             $ 400 \geq t $         & 0.2073 & \textbf{0.5009} & \textbf{0.5541} & \textbf{0.5981} & \textbf{0.6159}  \\   
        \end{tabular}
        \vspace{-15pt} 
    \end{table}
\end{center}
As seen here, our model is able to correctly predict at least one of the target actions as one of the 1-10 most likely choices between 24.7\% and 57\% of the time. We observe a 479.8\% increase in the top-1 accuracy when moving from mean to min, averaged across context lengths. This reduces to a 60.7\% increase for k = 5 and 33.6\% increase for k = 10. Larger models trained on more data are likely to push these bounds significantly higher and we are actively pursuing more sophisticated architectures to make use of the health-data specific embedding strategy we outline in section 3.
\\
\\
An important feature of our proposed architecture is the ability to restrain model output as a function of this measure of model certainty, i.e. higher learned separation proxies higher model certainty, and we demonstrate, in the table below, the degree to which conditioning on action-level certainty impacts predictive accuracy. Here we consider the mean-top-k accuracy over predictions where the mean separation over the target set $A'$ is above a given quantile. As such we are iteratively removing actions where the model is worse in differentiating between treatment and control. We are able to achieve remarkably high top-5 and top-10 accuracies for the top 20\% of actions in $\mathcal{A}$ as judged by their learned separation, reaching $\boldsymbol{99.26\%}$ for $k = 10$ and $\boldsymbol{50.74\%}$ for $k = 5$. Notably context length seems a more important predictor of model performance for $k = 1$ than mean separation.
\begin{center}
    \begin{table}[h]
    \centering
        \vspace{-5pt} 
        \caption{Mean-top-$k$ Accuracy for next-action prediction} 
        \begin{tabular}{l l l l l l } 
             \hline
              $\Delta^{\pi_n}(a)$ &  \multicolumn{5}{c}{Accuracy}  \\
              Quantile & k = 1 & k = 5 & k = 10 &  k = 15 & k = 20  \\ [0.5ex] 
             \hline\hline
                $Q$1 & 0.0147 & 0.1097 & 0.2273 & 0.2383 & 0.2582    \\ 
                $Q$2 & 0.0168 & 0.1248 & 0.2585 & 0.2710 & 0.2937    \\
                $Q$3 & 0.0192 & 0.1427 & 0.2956 & 0.3098 & 0.3357    \\
                $Q$4 & 0.0224 & 0.1666 & 0.3452 & 0.3616 & 0.3908    \\
                $Q$5 & 0.0268 & 0.1978 & 0.4065 & 0.4262 & 0.4611    \\
                $Q$6 & 0.0336 & 0.2475 & 0.5067 & 0.5275 & 0.5707    \\
                $Q$7 & 0.0447 & 0.3262 & 0.6687 & 0.6935 & 0.7384    \\
                $Q$8 & 0.0654 & 0.4710 & 0.9234 & 0.9293 & 0.9409    \\
                $Q$9 & \textbf{0.0705} & \textbf{0.5074} & \textbf{0.9926} & \textbf{0.9971} & \textbf{0.9975}    \\
        \end{tabular}
    \vspace{-15pt} 
    \end{table}
\end{center}

\subsubsection{Q-Accuracy}
In this section we consider the the empirical quantile function of an action $a$ in the predicted distribution of $\boldsymbol{\pi}^{\text{pre}}_n$ as a measure of accuracy. We construct this by estimating the probability that an action $a$ would have received lower probability under the estimator; 
\begin{equation}
    q\text{-accuracy}(a\;|\:I) \equiv 1 - \frac{\text{loc}[a \; | \; I]}{|\mathcal{A}|}
\end{equation}
When $q\text{-accuracy} = 1$ the location of $a$ is 0, i.e. the correct action receives the highest probability, and when $q\text{-accuracy} = 0$ the model is predicting the correct action last. Our model achieves a mean q-accuracy of \textbf{83.22\%} and median of \textbf{89.23\%} across a holdout sample of 300,000 action sequence pairs. That means that the median action $a \in A'$ receives higher probability than approximately 90\% of the possible action space. 
\\
\\
For each of the 300,000 order set pairs we take the mean q-accuracy over the actions in $A'$ as well as the mean order frequency and mean value of $\Delta^{\pi_n}(a)$. We then construct groups based on each of these three features as seen in Table 4 and compute the mean q-accuracy within each group. This gives us a sense of how important each of these features are in driving this measure of accuracy.
\begin{center}
    \begin{table}[h]
    \vspace{-5pt} 
    \centering
        \caption{Combined Q-Accuracy Measures} 
        \begin{tabular}{l l | l l | l l } 
             \hline
             Context Length & q-acc. & Order Freq. & q-acc. & $\Delta^{\pi_n}(a)$-quant. & q-acc. \\
             \hline\hline
              $ t < 50 $ & 0.8680 & Q1 &0.8038 & Q1& 0.6231  \\
              $ 50 \geq t < 250 $ & 0.8680 & Q2 &0.8906 &Q2 & 0.8277  \\
              $ 250 \geq t < 450 $ & 0.8996 & Q3&0.8635 & Q3& 0.9016  \\
              $ t > 450 $ & 0.9267 & Q4 &0.9413 & Q4&  0.9812 \\
        \end{tabular}
    \vspace{-5pt} 
    \end{table}
\end{center}
Again, $\Delta^{\boldsymbol{\pi}}$ is the most predictive feature for q-accuracy as conditioning on quantiles of action level mean separation leads to range of over 30\%-points. In the figure below we plot the full behavior of q-accuracy as a function of context length and log mean separation. Note that we take the log of $\Delta^\pi$ since this feature displays wide spread and, as seen below, a log transform establishes an approximately linear relationship. 
\begin{center}
    \includegraphics[scale = 0.5]{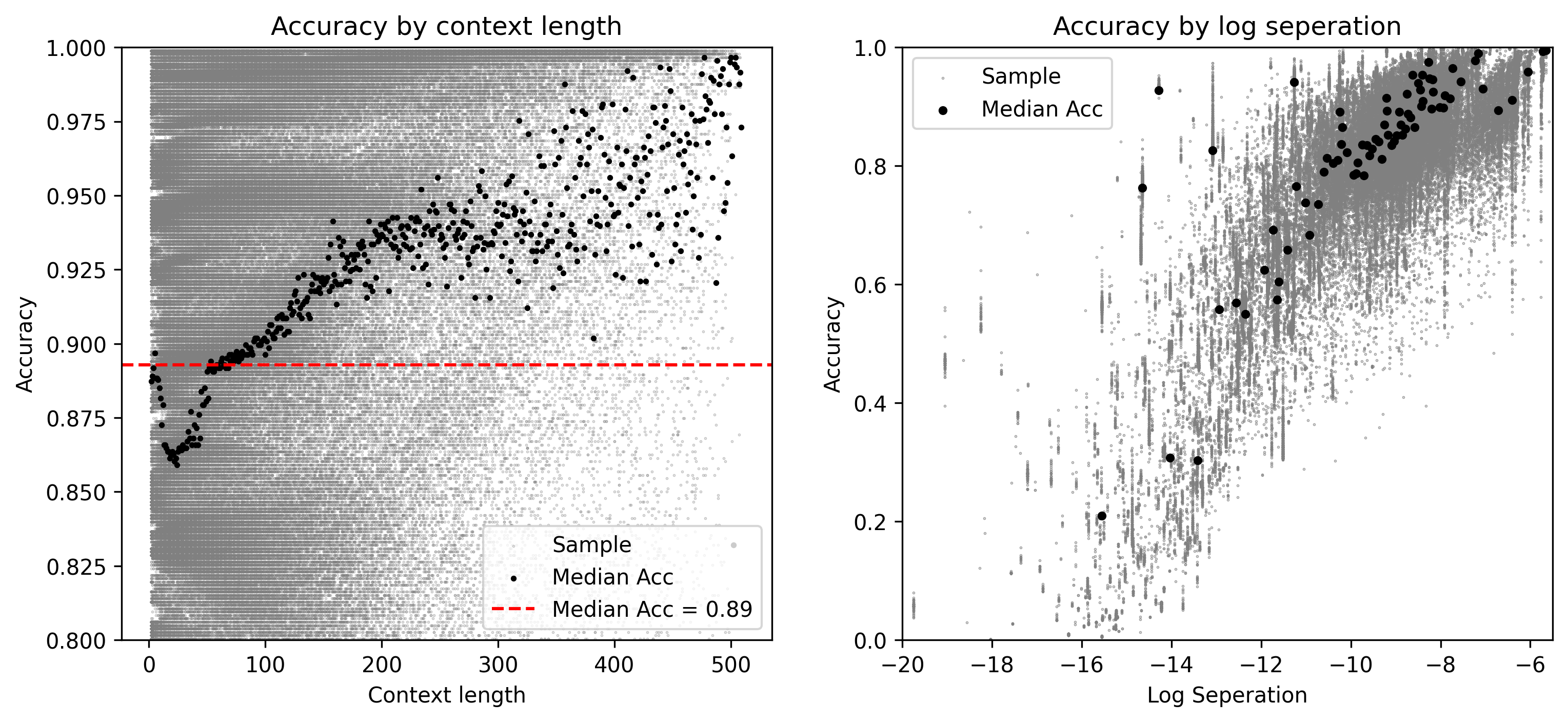}
    \textbf{Figure 2:} $q$-accuracy by context window and mean-$\log (\Delta^\pi)$ 
\end{center}
In gray we plot the 300,000 individual prediction instances in which multiple actions were predicted by the model. In black when then consider the median accuracy of all predictions made at a particular context length and quantile of $\Delta^\pi$. In both figures we see a strong positive correlation with diminishing returns.
\\
\\
The strong positive correlation between $t$ and accuracy is not surprising since predictions further along a patients path are likely more ``deterministic", i.e. more information is available to cut down the space of possible next actions. However, note also that the space of possible inputs increases massively with context length. For example, at a window length of 100, with $|\mathcal{A}| = 882$, there are approximately $1.102 \times 10^{134}$ possible information set configurations. Notably the space of observed realized paths decreases whilst the complexity of the input space grows massively. As such, the fact that accuracy increases to above 90\% for long context windows is a testament to the fact that even this very simple BPL-estimator is able to extract useful signal from increasingly complex and noisy inputs. A second interesting feature is the steep decrease in accuracy when context size increases from $t \geq 5$. Our hypothesis is that there exists a trade-off between the growing complexity of the input space and the decreasing fraction of realized information sets which leads to this sudden drop and later recovery. 
\\
\\
On the right hand side plot we repeat the analysis for variation in the mean learned separation of actions in the target set $A'$. Here we plot the log of the mean value of $\Delta^\pi$ over the actions in $A'$ against the achieved q-accuracy. This plot makes clear the strong positive relationship between learned separation over the actions in the target set and the ability of the model to correctly predict treatments. 

\section{Discussion}

Deep Causal Behavioral Policy Learning integrates methods and theory from multiple disciplines to model provider decision-making in response to complex clinical information, identify high-quality providers given patient characteristics, and learn the complex clinical practice patterns of these top providers. A proof-of-concept analysis illustrates the performance of a simple deep behavioral policy estimator on real-world clinical data from the UCSF emergency department (\cite{IC_CDW}). We pre-train a unimodal transformer using a next clinical action prediction task on 180,000 encounters from an emergency department, and evaluate the model using two primary metrics, mean top-k and quantile accuracy. As expected, accuracy increases dramatically with increasing context and data support. We further propose an action-level difference, ``learned separation",  as a measure of model certainty, and show that both top-k as well as quantile-based accuracy measures improve greatly when conditioning on learned separation. The same is true for for context length and order-frequency and we show that accuracy increases dramatically when more context is available and the target actions are more frequently featured in the training data. We believe this kind of analysis is crucial to deploying deep-learning technology safely in any medical domain. 
\\
\\
Our methodology provides a general approach with a range of practical applications, including causally rigorous quality measurement, provider coaching, and causally-grounded clinical decision support for complex longitudinal care. Contrasting observed patient outcomes with expected  counterfactual outcomes under assignment to a top provider serves as a metric for causal quality measurement, quantifying the potential for improvement in patient outcomes achievable by leveraging best practice patterns. The learned clinical behavioral policy of the optimal provider could also serve as the basis for provider coaching and up-skilling, as well as the basis for clinical decision support more broadly. Additional assumptions (such as robust transportability over time and place) would be needed to support such deployments. 
\\
\\
While the results of the empirical proof-of-concept analysis  we present are promising, training of the Large Clinical Behavioral Model on much larger comprehensive clinical databases, including complex patient state-spaces, and using more sophisticated architectures, is ongoing. For example, we currently predict the next action set recursively while prior work has demonstrated the utility of simple binary-prediction for this multi-label problem (\cite{CLIMBR}); we plan to build on this to construct a transformer architecture that can predict over the entire set of next actions efficiently. We are further working to expand our architecture to include both image and text data in the information-set, for which we are exploring a MoE-style architecture, as well as implementing the state-space embedding strategy we outline in section 3. In addition to architectural changes we are also working on additional evaluation approaches for models of this kind, especially focused on safe deployment. 
\\
\\
There are several limitations to the proposed methodology. First, our approach is premised on the existence of variation in provider practice patterns that affects patient outcomes. It further requires sufficient data support to both identify the optimal provider for a given set of patient characteristics and effectively fine-tune a general provider-agnostic (pre-trained) large clinical behavioral model to the clinical patterns of specific providers. In real-world applications, individual provider-level support is likely to limit the set of candidate providers considered, or require a coarsening beyond specific individual providers to provider types. Second, we make the simplifying assumption that a single provider is responsible for the majority of clinical decisions from the beginning of an encounter until the outcome was measured, but this is unlikely to hold in many healthcare settings where patients are seen and treated by multiple doctors, even within a single encounter such as a visit to the emergency department. Approaches to address this challenge are the topic of ongoing work. Finally, causal identification relies on the assumption that providers can serve as (conditional) instruments; this is reasonable in some, but not all clinical settings. In settings where quasi-random assignment of providers is unrealistic, alternative approaches to identifying provider-level effects may be required. Furthermore, behavioral policies trained on the observed clinical actions of providers may not reflect important unmeasured characteristics of care that affect outcomes (such as the quality of interpersonal interactions and the provider-patient relationship). As the comprehensiveness of multimodal data measured in the course of clinical interactions increases, these data can be incorporated into estimates of clinical behavioral policies.      
\\
\\
The core methods presented here suggest a number of interesting extensions. First, Deep Causal Behavioral Policy Learning, as described here, identifies the causal effects of provider-specific longitudinal clinical behavioral policies, rather than the causal effects of specific clinical actions. However, contrasting characteristics of the learned optimal behavioral policy with observed clinical behavioral policies also provides an opportunity to explore which policy characteristics causally affect expected patient outcomes. In ongoing work, we  develop approaches to quantify not only which behavioral policies result in improved patient outcomes and by how much (the focus of the current paper), but also which actions drive these differences. We can leverage the methods presented here to discover which lower-dimensional features of a given provider's complex clinical action mechanism are causal drivers of patient outcomes. Informally, this allows us to move from asking ``how would an optimal provider have behaved for a patient like this?" and ``how much would this behavior have changed outcomes" to ``what are the key clinical decisions that resulted in these improved outcomes?". 
\\
\\
An additional particularly interesting application of the Large Clinical Behavioral Model (in both its pre-trained provider-agnostic form, and after fine-tuning to the behavioral policies of optimal providers) is in training clinical reasoning models. Our conjecture is that our LCBM, by learning to represent the complex distribution of clinical paths, captures the underlying real world clinical logic of high quality providers. This is something which other models, for example LLMs trained on medical texts, are sorely missing, and which a behavioral policy could supply at scale. By leveraging the optimal behavioral policy as a reward model (i.e. a process reward as in \cite{Zhang_2025_ProcessRewardLessons} \cite{ Lightman_2023_StepByStep}, \cite{Li_2023_StepAwareVerifier}, \cite{ Uesato_2022_solvingmathwordproblems}, and \cite{Zhang_2025_ProcessRewardLessons}) one could align next-generation reasoning models with the underlying high-quality clinical logic embedded in our behavioral model. We turn to this in future work but mention it here since the causal aspect is crucial to making sure one is embedding identifiably \textit{high-quality} clinical logic into any reasoning model aimed at deployment in real-world clinical settings. 
\\
\\
In summary, in this paper, we present a deep learning approach to learn the actions of providers and causally identify high-quality decision-making. Our approach may be used in numerous clinical applications from decision support to quality measurement.  Extensions to this work are ongoing, and could enhance our understanding of what high-quality health care is and integrate this knowledge with modern day reasoning models. This paper offers an exciting and innovative new approach to measure and promote quality in today's healthcare system. 

\section{Acknowledgment}

The authors acknowledge the use of the UCSF Information Commons computational research platform, developed and supported by UCSF Bakar Computational Health Sciences Institute in collaboration with IT Academic Research Services, Center for Intelligent Imaging Computational Core, and CTSI Research Technology Program. The authors thank the Center for Healthcare Marketplace Innovation at UC Berkeley for support. The contents of this paper are subject to a patent application and covered under a patent filing. 

\newpage
\bibliography{bibliography}

\newpage

\end{document}